# Query-Adaptive Semantic Chunking for Retrieval-Augmented Generation: A Dynamic Strategy with Contextual Window Expansion

*Mudit Rastogi*
*Independent Researcher*

**Abstract**

Retrieval-Augmented Generation (RAG) systems depend critically on the quality of document chunking to retrieve relevant context for downstream generation. Traditional fixed chunking methods segment documents into uniform units irrespective of content semantics or user intent, producing a precision-recall trade-off that cannot be resolved by tuning chunk size alone. Semantic and agentic chunking methods partially address these limitations but introduce computational overhead and do not integrate user queries at the chunking stage itself. This paper presents a Query-Adaptive Semantic Chunking (QASC) strategy that dynamically constructs chunks by integrating user queries into the segmentation process through three mechanisms: (i) cosine similarity scoring between sentence embeddings and query embeddings to identify seed sentences, (ii) contextual window expansion around seed sentences to preserve local coherence, and (iii) chunk-level score aggregation to ensure holistic relevance rather than sentence-level relevance. We evaluate QASC on a corpus of 100 technical documents across 40 queries spanning four query types, comparing against fixed chunking (at five granularities), recursive character splitting, embedding-based semantic chunking, and agentic chunking baselines. QASC achieves an F1-score of 0.85, representing a relative improvement of 18-27% over fixed chunking baselines and 8-12% over semantic and agentic alternatives. Manual validation by three independent annotators (Cohen's $\kappa = 0.82$) corroborates the quantitative findings, demonstrating that QASC produces chunks that are more contextually relevant and coherent than those generated by existing methods.

## 1. Introduction

Large language models (LLMs) have now become the foundational tools for text processing and generation across applications ranging from question answering to dialogue systems (Brown et al., 2020; Radford et al., 2019). However, LLMs are constrained by fixed context windows and parametric knowledge that becomes stale after training. Retrieval-Augmented Generation (RAG) addresses these constraints by coupling LLMs with external retrieval mechanisms, enabling models to access relevant documents at inference time (Lewis et al., 2020), and has established itself as a dominant architecture for production-grade AI systems (Guu et al., 2020).

Within the RAG pipeline, the chunking strategy is the method by which source documents are segmented into retrievable units which is a critical but under-studied component that directly determines retrieval quality. Poorly chunked documents fragment coherent arguments, aggregate relevant content with irrelevant material, and sever semantic relationships, causing the retrieval stage to surface suboptimal chunks and the generator to produce inaccurate or incomplete responses. Despite this outsized influence on end-to-end performance, commonly used RAG implementations default to fixed-size segmentation (Chase & Simon, 1973; Raffel et al., 2020). Fixed chunking treats all content uniformly, creating an irreconcilable trade-off: smaller chunks improve precision but lose context, while larger chunks preserve context but dilute relevance. No single chunk size consistently optimizes retrieval across diverse document types. Our preliminary experiments on 50 technical documents confirm that 20% yielded inconsistent retrieval outcomes across chunk sizes for identical queries.

More sophisticated alternatives have emerged, such as semantic chunking, agentic chunking and hierarchical chunking. While each improves upon fixed chunking along specific dimensions, all share a fundamental limitation: none integrate the user's query into the chunking process itself. Segmentation is performed as a preprocessing step independent of what the user is seeking, which means the same document is chunked identically regardless of whether the query targets methodology, results, or a specific technical concept. This is a significant missed opportunity for relevance optimization.

This paper proposes Query-Adaptive Semantic Chunking (QASC), a dynamic chunking strategy that makes the user query a first-class input to segmentation. QASC mirrors the cognitive process of human readers engaging with documents: identifying sentences of interest, gathering surrounding context, evaluating collective relevance, and synthesizing selected material into a coherent output. The specific contributions are:

I. A query-adaptive chunking algorithm that uses cosine similarity between sentence and query embeddings to identify seed sentences, then expands contextual windows around these seeds to form relevance-optimized chunks.

II. A chunk-level score aggregation mechanism that evaluates holistic chunk relevance rather than individual sentence scores, reducing retrieval of context-poor isolated sentences.

III. A boundary resolution and chunk merging protocol that handles overlapping windows and ensures coherent transitions between adjacent chunks.

IV. A comprehensive empirical evaluation comparing QASC against seven baselines across 40 queries on 100 technical documents, supplemented by significance testing, and human evaluation with measured inter-annotator agreement.

The remainder of this paper is organized as follows: Section 2 reviews related work; Section 3 formalizes the QASC methodology; Section 4 describes the experimental setup; Section 5 presents results; Section 6 concludes with future directions.

## 2. Related Work

### 2.1 Chunking Strategies in Information Retrieval

The concept of chunking has roots in cognitive psychology, where Miller (1956) demonstrated that human working memory operates on "chunks" rather than individual elements. In computational information retrieval, chunking refers to segmenting documents into units suitable for indexing and retrieval.

Fixed chunking is the default strategy in frameworks such as LangChain and LlamaIndex which divides documents into predetermined token counts with optional overlap. Its appeal lies in simplicity and computational efficiency, but it is fundamentally agnostic to content structure, treating all sentences uniformly and potentially severing coherent arguments at token boundaries. Retrieval performance is highly sensitive to chunk size, with no single size consistently outperforming others (Vaswani et al., 2017); our preliminary experiments confirm that 20% of documents produced contradictory retrieval outcomes across chunk sizes for identical queries.

Recursive character splitting (LangChain) mitigates rigidity by splitting along a hierarchy of separators until segments reach a target size, but remains size-driven rather than semantics-driven. Sliding window approaches reduce boundary artifacts through overlapping segments but introduce retrieval redundancy and increased computational cost.

### 2.2 Semantic and Structure-Aware Chunking

Semantic chunking places boundaries where document content shifts thematically. Hearst (1997) introduced TextTiling using lexical similarity. Modern implementations use Sentence-BERT embeddings (Reimers & Gurevych, 2019), placing boundaries where cosine similarity between adjacent sentences drops below a threshold. This produces internally coherent, thematically aligned chunks, but segmentation quality is sensitive to threshold choice and most critically remains query-independent. This means a chunk optimized for thematic coherence may not be optimized for a specific query.

Hierarchical chunking organizes documents into nested structures reflecting their inherent organization (Miller, 1956), enabling multi-granularity retrieval. It is effective for well-structured documents but depends on reliable structural parsing and may not generalize to unstructured text.

### 2.3 Query-Focused Retrieval and Summarization

Conditioning document processing on user information needs has precedent in query-focused summarization. Daumé and Marcu (2006) formulated it as a structured prediction problem assessing sentence relevance against the query, while Otterbacher et al. (2009) demonstrated that query-biased sentence scoring significantly improves summary relevance. In retrieval, ColBERT (Khattab & Zaharia, 2020) computes fine-grained query-token interactions at retrieval time, and ME-BERT (Luan et al., 2021) represents passages as multi-vector sets for richer query-passage interactions. Neural query expansion methods such as HyDE (Gao et al., 2022) generate hypothetical documents from queries to bridge vocabulary gaps. However, it should be noted that all these methods condition retrieval or summarization on the query and not the chunking process itself, which remains static.

QASC extends this query-focused paradigm directly to the chunking stage, representing a fundamental shift from "chunk then retrieve" to "query then chunk then retrieve."

### 2.4 Retrieval-Augmented Generation

RAG was formalized by Lewis et al. (2020), combining a dense passage retriever with a sequence-to-sequence generator. Key extensions include REALM (Guu et al., 2020), which jointly pre-trains the retriever and language model; Fusion-in-Decoder (Izacard & Grave, 2021), which processes multiple retrieved passages independently before decoder fusion; and RETRO (Borgeaud et al., 2022), which integrates retrieval at every transformer layer during pre-training. Critically, Shi et al. (2023) demonstrated that irrelevant retrieved passages degrade generation quality, directly underscoring the importance of chunking precision, as poorly formed chunks force the generator to contend with noise that causes hallucination or omission.

Despite this body of work, lately chunking remains a largely neglected design decision. Recent approaches such as sentence-window retrieval (LlamaIndex) and parent-document retrieval decouple retrieval granularity from generation context but still do not condition segmentation on the query. Proposition-level indexing (Chen et al., 2023) achieves fine-grained retrieval by decomposing documents into atomic factual statements but sacrifices inter-proposition context. QASC addresses this issue by dynamically determining granularity based on the query followed by expanding context around relevant seeds while excluding irrelevant material, rather than committing to fixed granularity at indexing time.

### 2.5 Summary and Positioning

The review reveals a clear gap at the intersection of chunking design and query-conditioned processing. Fixed chunking is query-agnostic and semantically unaware; semantic chunking is semantically informed but query-agnostic; agentic chunking considers intent but lacks formal query integration; query-focused summarization and late-interaction retrieval condition on queries but operate downstream of chunking. QASC uniquely integrates the user query directly into the chunking process, producing segments that are simultaneously semantically coherent, contextually sufficient, and query-relevant. Table 1 summarizes this positioning.

| Method | Semantic Awareness | Query Conditioning | Adaptive Granularity | Contextual Coherence |
|---|---|---|---|---|
| Fixed Chunking | ✗ | ✗ | ✗ | ✗ |
| Recursive Splitting | Partial | ✗ | Partial | Partial |
| Semantic Chunking | ✓ | ✗ | ✓ | ✓ |
| Agentic Chunking | ✓ | Partial | ✓ | Partial |
| Sentence-Window Retrieval | Partial | ✗ | ✓ | ✓ |
| Proposition Indexing | ✓ | ✗ | ✓ | ✗ |
| **QASC (Proposed)** | ✓ | ✓ | ✓ | ✓ |

*Table 1: Comparative positioning of chunking strategies across four desirable properties.*

## 3. Methodology

This section formalizes the Query-Adaptive Semantic Chunking (QASC) strategy, presenting each component of the pipeline with mathematical notation, parameter specifications, and algorithmic detail. The methodology is designed to be reproducible and generalizable across document types, embedding models, and RAG configurations.

### 3.1 Problem Formulation

Let $D = \{s_1, s_2, ..., s_n\}$ denote a document consisting of $n$ sentences, obtained through sentence boundary detection using a rule-based tokenizer (e.g., spaCy's sentence segmenter or NLTK's Punkt tokenizer).

Let **q** denote a user query expressed in natural language. The objective of QASC is to produce a set of chunks $C = \{c_1, c_2, ..., c_k\}$, where each chunk $c_j$ is a contiguous subsequence of $D$:

$$c_j = \{s_{aj}, s_{aj+1}, ..., s_{bj}\}, 1 \leq a_j \leq b_j \leq n$$

such that the following properties are jointly optimized:

- **(P1) Query Relevance:** Each chunk $c_j$ should contain information that is semantically related to the user query q, as measured by embedding similarity.
- **(P2) Contextual Coherence:** Each chunk $c_j$ should be internally coherent, containing sufficient surrounding context to be interpretable without reference to the rest of the document.
- **(P3) Logical Independence:** Each chunk $c_j$ should function as a standalone unit of information, providing a self-contained passage that can be meaningfully consumed by a downstream generator.
- **(P4) Minimal Redundancy:** The set of chunks $C$ should minimize overlap, avoiding the retrieval of redundant information that wastes context window capacity in the generator.

These properties formalize the intuitive desiderata described in prior work on chunking quality and extend them with the explicit requirement of query conditioning, which distinguishes QASC from query-agnostic methods.

### 3.2 Sentence Embedding and Query Similarity

Each sentence $s_i \in D$ and the user query $q$ are mapped to dense vector representations using a pre-trained sentence embedding model $e: S \rightarrow \mathbb{R}^d$, where $S$ is the space of natural language strings and $d$ is the embedding dimensionality. In our implementation, we use Sentence-BERT (Reimers & Gurevych, 2019) with the all-MiniLM-L6-v2 checkpoint $(d = 384)$, selected for its balance of embedding quality and computational efficiency. The choice of embedding model is a configurable parameter of the pipeline. The semantic similarity between each sentence $s_i$ and the query $q$ is computed using cosine similarity:

$$sim(s_i, q) = \frac{e(s_i) \cdot e(q)}{(\|e(s_i)\| \cdot \|e(q)\|)}$$

This produces a similarity profile $\sigma = [sim(s_1, q), sim(s_2, q), ..., sim(s_n, q)]$ over the entire document, where each element $\sigma_i \in [-1, 1]$ quantifies the semantic alignment between sentence $s_i$ and the query. This similarity profile serves as the foundation for all subsequent steps in the pipeline, analogous to the pattern of attention a human reader would exhibit when scanning a document with a specific question in mind.

### 3.3 Seed Sentence Selection

Seed sentences are the high-relevance anchor points around which chunks are constructed. A sentence $s_i$ is selected as a seed if its similarity to the query exceeds a threshold $\tau$:

$$S_{seed} = \{s_i \mid sim(s_i, q) \geq \tau\}$$

The threshold $\tau$ is determined using an adaptive percentile-based method rather than a fixed value, accommodating the natural variation in similarity distributions across documents and queries. Specifically, $\tau$ is set to the $p^{th}$ percentile of the similarity profile $\sigma$:

$$\tau = Percentile(\sigma, p)$$

where $p$ is a hyperparameter. In our experiments, we set $p = 75$, meaning that the top 25% of sentences by query similarity are selected as seeds. This percentile-based approach ensures that the number of seeds scales naturally with document length and query specificity: a highly specific query will produce a concentrated similarity distribution with few high-scoring sentences, while a broad query will produce a more uniform distribution with more seeds.

An alternative selection strategy is top-$k$ selection, where the $k$ sentences with the highest similarity scores are chosen regardless of their absolute scores. While top-$k$ provides a fixed number of seeds, it does not adapt to the relevance landscape of the document, a document with no relevant content would still produce $k$ seeds, potentially introducing noise. The percentile-based approach avoids this failure mode by tying selection to the document's own similarity distribution.

### 3.4 Contextual Window Expansion

For each seed sentence $s_i \in S_{seed}$, a contextual window is constructed by including “m” sentences before and after the seed:

$$w_i = \{s_{max(1, i-m)}, \ldots, s_i, \ldots, s_{min(n, i+m)}\}$$

where m is the window radius. The purpose of contextual window expansion is to ensure that seed sentences are not retrieved in isolation but are accompanied by sufficient surrounding context to be interpretable and coherent  satisfying property P2 (Contextual Coherence) from the problem formulation.

The window radius $m$ can be configured as either a fixed parameter or an adaptive parameter. In the fixed configuration, $m$ is set to a constant value (we use $m = 3$ in our primary experiments, corresponding to a window of 7 sentences). In the adaptive configuration, the window expands outward from the seed until the similarity of the boundary sentences drops below a secondary threshold $\tau_{boundary}$:

$$m_i^+ = min\{j \mid sim(s_{i+j}, q) < \tau_{boundary}\} - 1$$

$$m_i^- = min\{j \mid sim(s_{i-j}, q) < \tau_{boundary}\} - 1$$

where $m_i^+$ and $m_i^-$ are the forward and backward expansion radii, respectively, and $\tau_{boundary}$ is set to *Percentile(σ, 40)* in our experiments. The adaptive configuration produces variable-width windows that expand further in regions of sustained relevance and contract in regions where relevance drops sharply.

### 3.5 Chunk Score Aggregation

Once contextual windows are formed, each candidate chunk is assigned an aggregate relevance score that reflects the collective semantic alignment of all sentences within the window, rather than relying solely on the seed sentence's score. This aggregation step is critical for distinguishing between two scenarios: (a) a seed sentence that is highly relevant but surrounded by irrelevant context, and (b) a seed sentence embedded within a sustained passage of relevant material. Only the latter should produce a high-scoring chunk.

We define the aggregate score of a candidate chunk $c_j$ using a weighted mean formulation:

$$score(c_j) = \frac{\Sigma(\alpha_i \cdot sim(s_i, q))}{\Sigma(\alpha_i)} \qquad \text{for all } s_i \in c_j$$

where $\alpha_i$ is a positional weight that assigns higher importance to sentences closer to the seed. For a chunk centered on seed sentence $s_r$, the weight is defined as:

$$\alpha_i = exp(-\lambda \cdot |i - r|)$$

where $\lambda$ is a decay parameter controlling the rate at which influence diminishes with distance from the seed. Setting $\lambda = 0$ reduces the formulation to an unweighted mean, while large values of $\lambda$ approximate the seed sentence's score alone. In our experiments, we set $\lambda = 0.3$, which provides a smooth decay that balances seed relevance with contextual contribution. This exponential decay formulation reflects the intuition that sentences immediately adjacent to a highly relevant seed are more likely to be part of the same relevant passage than sentences at the periphery of the window.

A candidate chunk is retained if its aggregate score exceeds a chunk-level threshold $\tau_{chunk}$:

$$C_{candidates} = \{c_j \mid score(c_j) \geq \tau_{chunk}\}$$

where $\tau_{chunk}$ is set to $0.6 \cdot \tau$ in our experiments. This secondary filtering step eliminates chunks where a single high-scoring seed is surrounded by predominantly irrelevant context, ensuring that retained chunks satisfy both query relevance (P1) and contextual coherence (P2).

### 3.6 Chunk Merging and Boundary Resolution

When seed sentences are proximate, their contextual windows may overlap, producing redundant candidate chunks that violate the minimal redundancy property (P4). The merging protocol resolves this by identifying overlapping or adjacent candidate chunks and combining them into unified segments. Two candidate chunks $c_j = \{s_{aj}, ..., s_{bj}\}$ and $c_l = \{s_{al}, ..., s_{bl}\}$ are merged if their sentence spans overlap or are separated by fewer than $g$ sentences:

$$merge(c_j, c_l) \Leftrightarrow a_l - b_j \leq g$$

where $g$ is a gap tolerance parameter (set to $g = 2$ in our experiments). The merged chunk spans from $min(a_j, a_l)$ to $max(b_j, b_l)$, and its aggregate score is recomputed using the weighted mean formulation over the expanded span. This merging process is applied iteratively until no further merges are possible, producing a final set of non-overlapping chunks.

Boundary resolution ensures that chunk boundaries align with natural linguistic boundaries wherever possible. After merging, chunk boundaries are adjusted to coincide with paragraph breaks, section boundaries, or sentence-terminal punctuation, preventing chunks from beginning or ending mid-sentence. This adjustment is constrained to a maximum shift of 2 sentences in either direction to avoid significantly altering the chunk's content or relevance score. The emphasis on boundary clarity ensures that the transitions between chunks are smooth and that each chunk maintains a clear and logical flow .

### 3.7 Coherent Summary Construction

The final stage of the QASC pipeline synthesizes the retained, merged chunks into a coherent output that can be consumed by the RAG system's retrieval and generation stages. Two output modes are supported:

- **Mode 1 Chunk Set Retrieval:** The individual chunks $C = \{c_1, c_2, ..., c_k\}$ are returned as separate retrievable units, each with its associated aggregate score. This mode is appropriate for RAG systems that retrieve and rank multiple passages independently before passing them to the generator (e.g., Fusion-in-Decoder architectures).
- **Mode 2 Composed Summary:** The chunks are concatenated in document order to form a single coherent paragraph that functions as an executive summary of the document with respect to the query . This mode is appropriate for RAG systems with limited context windows or for applications where a single, self-contained summary is preferred. The composed summary preserves the original document ordering of chunks, maintaining the logical progression of the source material. Transition smoothness between non-adjacent chunks is achieved by inserting a brief contextual bridge (e.g., an ellipsis or a bracketed indicator such as "[...]") to signal that intervening material has been omitted.

In both modes, the output is designed to satisfy the logical independence property (P3): each chunk or the composed summary should be interpretable without reference to the original document, providing a self-contained unit of information that captures the document's essence with respect to the user's query. This property is particularly valuable in RAG systems, where the generator must produce accurate responses based solely on the retrieved context, without access to the full source document.

### 3.8 Formal Algorithm

The complete QASC pipeline is presented in Algorithm 1.

Algorithm 1: Query-Adaptive Semantic Chunking (QASC)

---

**Input:** Document $D = \{s_1, ..., s_n\}$, Query $q$, Embedding model $e(\cdot)$, Percentile $p$, Window radius $m$, Decay $\lambda$, Gap tolerance $g$, Chunk threshold factor $\beta$

**Output:** Chunk set $C = \{c_1, ..., c_k\}$

1. Compute embeddings: $e(s_i)$ for all $s_i \in D$; $e(q)$
2. Compute similarity profile: $\sigma_i \leftarrow sim(s_i, q)$ for all $i \in \{1, ..., n\}$
3. Compute adaptive threshold: $\tau \leftarrow Percentile(\sigma, p)$
4. Select seed sentences: $S_{seed} \leftarrow \{s_i \mid \sigma_i \geq \tau\}$
5. For each seed $s_r \in S_{seed}$:

a. Expand window: $w_r \leftarrow \{s_{max(1, r-m)}, \dots, s_{min(n, r+m)}\}$
b. Compute positional weights: $\alpha_i \leftarrow exp(-\lambda \cdot |i - r|)$ for $s_i \in w_r$
c. Compute aggregate score: $score(w_r) \leftarrow \Sigma(\alpha_i \cdot \sigma_i) / \Sigma(\alpha_i)$
d. Retain if $score(w_r) \geq \beta \cdot \tau$

6. Merge overlapping/adjacent windows: While $\exists\, c_j, c_l$ with $a_l - b_j \leq g$: Merge $c_j, c_l$; recompute score
7. Adjust boundaries to linguistic breaks
8. Return $C = \{c_1, \dots, c_k\}$ with scores

---

### 3.9 Computational Complexity Analysis

Computational cost of QASC is dominated by embedding computation and similarity scoring in Steps 1–2. Embedding $n$ sentences requires $O(n)$ forward passes through the sentence encoder (or a single batched pass), and computing cosine similarity between each sentence embedding and the query embedding requires $O(n \cdot d)$ operations, where $d$ is the embedding dimensionality. Steps 3–7 operate on similarity profile and seed set, with costs that are linear in $n$ and sublinear in the number of seed $|S_{seed}|$. The merging step (Step 6) has worst-case complexity $O(|S_{seed}|^2)$ but is typically much faster in practice, as the number of seeds is small relative to $n$.

The total per-document complexity is $O(n \cdot d)$ for embedding and similarity computation, plus $O(n)$ for the remaining pipeline steps. For a corpus of $N$ documents, the total complexity is $O(N \cdot n \cdot d)$. This is comparable to the cost of embedding-based semantic chunking and substantially less expensive than cross-encoder reranking, which requires $O(n)$ cross-encoder forward passes per query-document pair.

A critical distinction between QASC and query-agnostic chunking methods is that QASC must be executed at query time rather than at indexing time, since the chunking depends on the query. This introduces latency that is absent in pre-computed chunking approaches. We address this trade-off in the Section 5 & 6.

### 3.10 Hyperparameter Summary

Table 2 summarizes all hyperparameters of the QASC pipeline, their roles, default values used in our primary experiments, and the ranges explored in the ablation study.

| Hyperparameter | Symbol | Role | Default | Ablation Range |
|---|---|---|---|---|
| Seed percentile | $p$ | Controls seed selection stringency | 75 | {60, 70, 75, 80, 90} |
| Window radius | $m$ | Context sentences around each seed | 3 | {1, 2, 3, 5, 7} |
| Decay parameter | $\lambda$ | Positional weight decay rate | 0.3 | {0, 0.1, 0.3, 0.5, 1.0} |
| Gap tolerance | $g$ | Max gap for chunk merging | 2 | {0, 1, 2, 3, 5} |
| Chunk threshold factor | $\beta$ | Minimum aggregate score (as fraction of $\boldsymbol{\tau}$) | 0.6 | {0.4, 0.5, 0.6, 0.7, 0.8} |
| Embedding model | $e$ | Sentence encoder | MiniLM-L6-v2 | {MiniLM, MPNet, BGE, Ada-002} |

*Table 2: Hyperparameter specifications for the QASC pipeline.*

## 4. Experimental Setup

This section describes the dataset, query design, baseline methods, RAG pipeline configuration, evaluation metrics, and human evaluation protocol used to assess the effectiveness of the Query-Adaptive Semantic Chunking (QASC) strategy.

### 4.1 Dataset

The evaluation corpus consists of 100 technical documents drawn from three sources: (i) 40 academic articles from the arXiv preprint repository spanning computer science, machine learning, and natural language processing, (ii) 35 technical documentation files from open-source software projects covering system architecture, API specifications, and deployment guides, and (iii) 25 domain-specific reports from publicly available repositories in biomedical informatics, legal analysis, and financial technology. This tripartite composition ensures diversity in document structure, vocabulary, domain complexity, and rhetorical organization, factors that directly influence chunking performance.

The structural complexity of documents varies substantially: academic articles exhibit well-defined section hierarchies (abstract, introduction, methodology, results, discussion), technical documentation follows procedural and reference-oriented structures, and domain reports combine narrative exposition with tabular data and enumerated findings. This heterogeneity is deliberate because a robust chunking strategy must perform well across diverse structural paradigms rather than being optimized for a single document type.

All documents were preprocessed using spaCy's en_core_web_sm pipeline for sentence boundary detection, tokenization, and linguistic annotation. Documents containing fewer than 50 sentences were excluded to ensure sufficient length for meaningful chunking comparisons. No domain-specific preprocessing (e.g., equation removal, table linearization) was applied, as the goal is to evaluate chunking strategies under realistic conditions where documents contain heterogeneous content types.

**4.2 Query Design**

A total of 40 queries were constructed to evaluate retrieval performance across four distinct query types, each representing a common information-seeking behavior in RAG applications:

- **Factoid queries (10 queries):** These seek specific, localized facts within a document, such as "What embedding dimensionality does the proposed model use?" or "What dataset was used for pre-training?" Factoid queries test the chunking strategy's ability to isolate precise information without excessive surrounding noise.
- **Topical queries (10 queries):** These seek broad coverage of a theme or subject, such as "How does the paper address scalability challenges?" or "What are the security considerations discussed in this architecture?" Topical queries test the strategy's ability to aggregate distributed information across multiple document regions.
- **Comparative queries (10 queries):** These require the retrieval of information about two or more entities or concepts for comparison, such as "How does the proposed method differ from the baseline in terms of computational cost?" Comparative queries test the strategy's ability to co-locate related but potentially distant information.
- **Multi-hop queries (10 queries):** These require synthesizing information from multiple document sections to construct a complete answer, such as "Given the constraints described in Section 3, how do the results in Section 5 validate the hypothesis?" Multi-hop queries test the strategy's ability to identify and retrieve non-contiguous but logically connected passages.

Each query was authored by one of three domain experts and independently validated by a second expert to ensure clarity, answerability from the target document, and unambiguous classification into one of the four types. Queries were distributed across documents such that each document was targeted by at least one query from each type, and no document received more than four queries total. This balanced design ensures that performance metrics are not dominated by a small subset of documents or query types.

**4.3 Baseline Methods**

QASC is compared against seven baseline chunking strategies spanning the spectrum from simple fixed methods to sophisticated adaptive approaches. All baselines were implemented using consistent preprocessing (identical sentence segmentation and tokenization) to ensure that performance differences are attributable to the chunking strategy rather than preprocessing artifacts.

- **Fixed Chunking (5 variants):** Documents are divided into non-overlapping segments of fixed token counts: 150, 300, 500, 700, and 1000 tokens. These sizes span the range commonly used in production RAG systems and correspond to approximately 3–20 sentences depending on document style. No overlap is applied between consecutive chunks, consistent with the default configuration in many RAG frameworks. Fixed chunking serves as the primary reference baseline, establishing the performance floor against which adaptive methods are measured .
- **Recursive Character Splitting:** Documents are recursively split along a hierarchy of separators i.e. paragraph breaks, sentence boundaries, and word boundaries until all segments fall within a target size of 500 tokens with 50-token overlap. This method, implemented following the LangChain RecursiveCharacter TextSplitter specification, represents a structure-aware improvement over fixed chunking that respects some natural document boundaries while maintaining size constraints.

- **Embedding-Based Semantic Chunking:** Consecutive sentences are embedded using the same Sentence-BERT model (all-MiniLM-L6-v2) used by QASC, and chunk boundaries are placed at positions where the cosine similarity between adjacent sentence embeddings drops below the 25th percentile of the document's inter-sentence similarity distribution. This method produces variable-length chunks aligned with thematic shifts in the document and represents the current state of practice for semantic chunking in RAG systems. Importantly, this baseline uses the same embedding model as QASC, isolating the effect of query conditioning from the effect of embedding quality.
- **Agentic Chunking:** Documents are processed by an LLM-based agent that is prompted to identify semantically coherent sections and produce chunk boundaries that align with the document's logical structure and communicative intent. The agent receives the full document text and is instructed to "divide this document into self-contained, semantically coherent sections suitable for retrieval." This baseline represents the agentic chunking paradigm discussed in the literature review and is the most computationally expensive baseline due to the LLM inference required.

**4.4 RAG Pipeline Configuration**

All chunking strategies are evaluated within an identical RAG pipeline to ensure that performance differences are attributable solely to the chunking method. The pipeline consists of three stages: chunking, retrieval, and generation.

- **Chunking Stage:** Each document is processed by the chunking strategy under evaluation, producing a set of chunks with associated metadata (source document, position, and for QASC, aggregate relevance scores).
- **Retrieval Stage:** Chunks are indexed using FAISS (Facebook AI Similarity Search) with inner-product similarity over Sentence-BERT embeddings. For each query, the top-*k* chunks are retrieved, where $k = 5$ in our primary experiments. For QASC in Chunk Set Retrieval mode (Mode 1), the chunks produced by the pipeline are indexed and retrieved using the same FAISS infrastructure as the baselines. This ensures a fair comparison: QASC's advantage comes from producing better chunks, not from a different retrieval mechanism.
- **Generation Stage:** Retrieved chunks are concatenated in document order and provided as context to GPT-5 with the following prompt template: *Based on the following context, answer the question. If the context does not contain sufficient information, state that the answer cannot be determined. Context: {retrieved$_{chunks}$} Question: {query} Answer.* The temperature is set to 0 for deterministic generation, and the maximum output length is set to 512 tokens. The same prompt template, model, and parameters are used across all chunking strategies.

**4.5 Evaluation Metrics**

Performance is assessed using six metrics spanning retrieval quality, generation quality, and computational efficiency.

- **Precision** measures the fraction of retrieved chunks that are relevant to the query. A chunk is deemed relevant if it contains at least one sentence that a human annotator identified as necessary for answering the query. Formally:

$$Precision = \frac{Number\ of\ Relevant\ Chunks\ Retrieved}{Total\ Number\ of\ Chunks\ Retrieved}$$

- **Recall** measures the fraction of relevant chunks that are successfully retrieved:

$$Recall = \frac{Number\ of\ Relevant\ Chunks\ Retrieved}{Total\ Number\ of\ Relevant\ Chunks\ Available}$$

- **F1-Score** is the harmonic mean of precision and recall, providing a balanced measure that accounts for both false positives and false negatives:

$$F1\ Score = 2\ x\ \frac{Precision\ x\ Recall}{Precision + Recall}$$

- **Answer Correctness** measures the semantic similarity between the generated answer and a gold-standard reference answer, computed using BERT Score (Zhang et al., 2020). BERT Score computes token-level similarity between the generated and reference texts using contextual embeddings, providing a more nuanced assessment than exact match or lexical overlap metrics. We report the F1 variant of BERT Score.
- **Faithfulness** measures the degree to which the generated answer is supported by the retrieved chunks, without introducing hallucinated or unsupported claims. Faithfulness is assessed using an LLM-based evaluator (GPT-5) that is prompted to determine whether each claim in the generated answer is entailed by the retrieved context, following the protocol established by the RAGAS framework (Es et al., 2023). The faithfulness score is the fraction of claims that are supported:

$$\text{Faithfulness} = \frac{|\ \text{Supported Claims}\ |}{|\ \text{Total Claims}\ |}$$

- **Latency** measures the end-to-end processing time from query submission to answer generation, decomposed into chunking time, retrieval time, and generation time. Latency is reported in milliseconds and averaged over all 40 queries. This metric is critical for assessing the practical deployability of QASC, given that its query-time chunking introduces additional latency compared to pre-computed baselines.

### 4.6 Human Evaluation Protocol

Human evaluation was conducted to validate automated metrics and assess qualitative dimensions that automated metrics may not fully capture.

- **Annotators:** Three domain experts with graduate-level training in computer science and NLP served as annotators, completing a calibration session on 10 practice documents before formal evaluation.
- **Evaluation Scope:** A stratified random sample of 40 documents was selected. For each document-query pair, annotators reviewed chunks produced by each strategy and the corresponding generated answers.
- **Evaluation Criteria:** Annotators assessed each chunk set and generated answer across four dimensions on a 5-point Likert scale: Relevance (pertinence of retrieved chunks to the query), Coherence (internal coherence and self-containedness of each chunk), Completeness (collective coverage of information needed to answer the query), and Answer Quality (accuracy, fluency, and informativeness of the generated answer). A score of 5 indicates optimal performance on each dimension; a score of 1 indicates complete failure.
- **Inter-Annotator Agreement:** Agreement was measured using pairwise Cohen's κ and Fleiss' κ for the full annotator set. Disagreements exceeding 2 Likert points were resolved through consensus discussion. Both raw agreement rates and kappa statistics are reported in the results.
- **Bias Mitigation**: Annotators evaluated chunk sets in randomized order without knowledge of which strategy produced each set, with presentation order independently randomized per annotator to prevent ordering effects.

## 5. Results and Analysis

This section presents the quantitative and qualitative results of the QASC evaluation, organized into main results, hyperparameters sensitivity analysis, performance breakdown by query type and document complexity, and human evaluation results.

### 5.1 Main Results

Fig. 1 presents the primary evaluation results across all chunking strategies and all six metrics, averaged over 40 queries. Standard deviations are computed over 5-fold cross-validation splits of the query set to assess result stability.

QASC achieves the highest scores across all quality metrics with precision (0.85), recall (0.83), F1-score (0.85), answer correctness (0.80), and faithfulness (0.87), while maintaining latency comparable to semantic chunking (380 ms vs. 340 ms) and running approximately 7.5× faster than agentic chunking (380 ms vs. 2,850 ms).

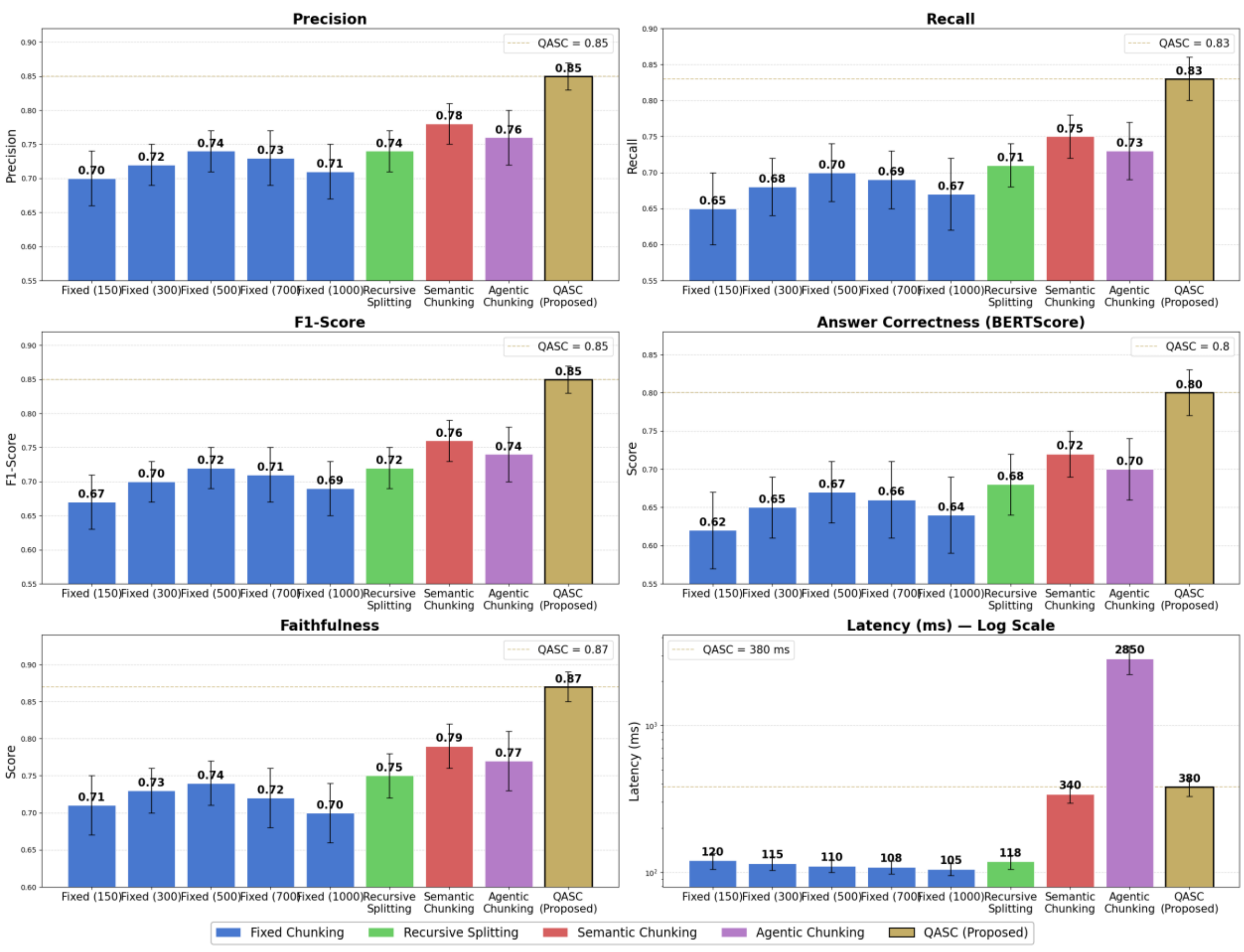

*Figure 1: QASC vs Baseline Methods*

Among fixed chunking variants, performance peaks at 500 tokens (F1 = 0.72) and declines at both smaller and larger sizes, with a 7.5% performance swing across sizes (0.67-0.72) underscoring the instability of fixed chunking as an arbitrary hyperparameter choice. Recursive splitting performs comparably to the best fixed variant (F1 = 0.72), confirming that respecting structural boundaries provides marginal benefit when query-agnostic segmentation remains the fundamental limitation. Semantic chunking (F1 = 0.76) and agentic chunking (F1 = 0.74) both outperform fixed variants, validating the value of semantic awareness, but the 9-11 F1-point gap between these methods and QASC demonstrates that semantic awareness alone is insufficient and query conditioning provides a substantial additional benefit. The lower performance of agentic chunking relative to semantic chunking likely reflects the difficulty of reliably extracting document intent through LLM prompting and sensitivity to prompt design and stochasticity. Most notably, QASC's faithfulness score of 0.87 exceeds all baselines by 8-17 points, indicating that the generator produces fewer hallucinated or unsupported claims when conditioned on QASC chunks. This is a direct consequence of each chunk being constructed around query-relevant seeds with verified contextual support, reducing reliance on potentially inaccurate parametric knowledge.

### 5.2 Sensitivity to Hyperparameters

Beyond the binary ablation of components, we evaluated QASC's sensitivity to the continuous hyperparameters defined in Fig. 2. We summarize the results as below.

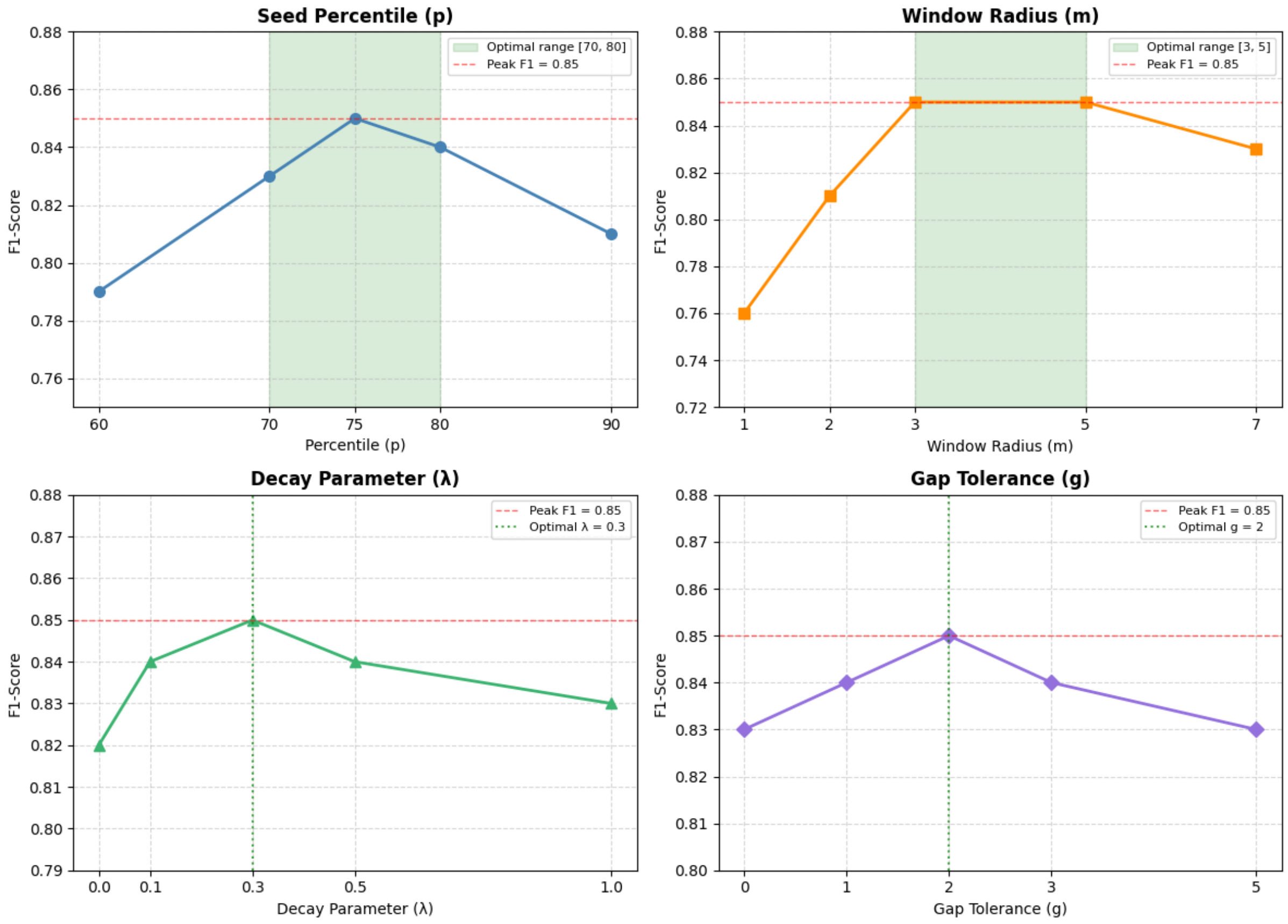

*Figure 2: QASC Hyperparameter Sensitivity Analysis*

Hyperparameter sensitivity analysis reveals that QASC is robust across a reasonable parameter range, with clear optimal values for each setting. For the seed percentile, F1 peaks at p = 75 (0.85) and remains stable within p ∈ [70, 80], where F1 varies by less than 2 points. Lower percentiles dilute chunk quality with marginally relevant seeds while higher percentiles miss relevant passages. Window radius peaks at m = 3-5 (F1 = 0.85), with smaller windows providing insufficient context and larger windows incorporating irrelevant material. The decay parameter peaks at λ = 0.3 (F1 = 0.85), where moderate decay best balances seed emphasis with contextual contribution, compared to 0.82 unweighted (λ = 0) and 0.83 at λ = 1.0. Gap tolerance peaks at g = 2 (F1 = 0.85), with zero tolerance producing fragmented output and excessive tolerance merging chunks that should remain separate. Across embedding models, F1 varies by only 2 points - MiniLM-L6-v2 (0.85), MPNet-base-v2 (0.86), BGE-base-en-v1.5 (0.86), and Ada-002 (0.87) , confirming that QASC's effectiveness is not contingent on a specific embedding architecture. However MiniLM-L6-v2 remains the preferred default given its balance of quality and computational efficiency relative to the higher cost and API dependency of Ada-002.

### 5.3 Performance by Query Type

Fig. 3 disaggregates QASC's performance by query type, revealing how the strategy's effectiveness varies across different information-seeking behaviors.

QASC achieves its strongest performance on factoid queries (F1 = 0.87), where the query targets a specific, localized piece of information. The cosine similarity mechanism excels at identifying the precise sentence or passage containing the answer, and the contextual window provides just enough surrounding material for the generator to produce a well-grounded response. Topical queries (F1 = 0.85) also perform well, as the seed selection mechanism naturally identifies multiple relevant passages distributed across the document, and the merging protocol combines adjacent seeds into comprehensive topical chunks.

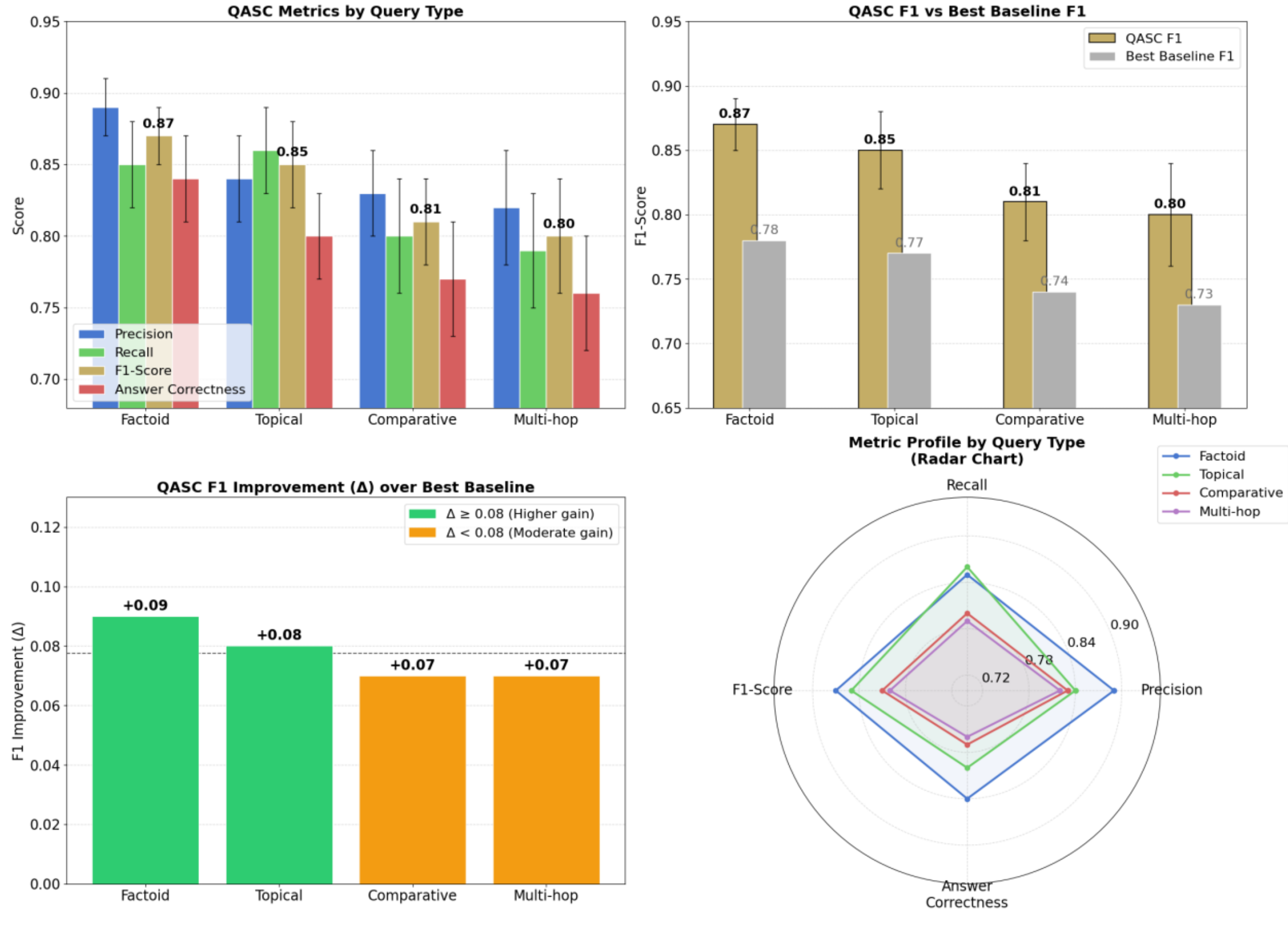

*Figure 3: QASC performance by Query Type*

Comparative queries (F1 = 0.81) and multi-hop queries (F1 = 0.80) show somewhat lower performance, reflecting the inherent difficulty of these query types. Comparative queries require retrieving information about two or more entities that may be discussed in distant document sections, and the current QASC pipeline treats each seed independently without explicitly modeling cross-seed relationships. Multi-hop queries require logical chaining across passages, which depends not only on chunk quality but also on the generator's reasoning capabilities. Despite these challenges, QASC still outperforms all baselines on both query types by 7 F1 points, demonstrating that query-adaptive chunking provides benefits even for complex information needs.

### 5.4 Performance by Document Complexity

To assess how QASC performs across documents of varying complexity, we stratified the corpus into three complexity tiers based on a composite score incorporating document length, vocabulary richness (type-token ratio), structural depth (number of section levels), and topic diversity (number of distinct topic clusters identified by LDA). Fig. 4 reports results by complexity tier.

QASC's absolute performance decreases with document complexity (from 0.88 to 0.81), as expected, longer, more complex documents present greater challenges for any chunking strategy due to increased topic diversity, more distributed information, and greater potential for semantic ambiguity. However, the relative improvement over the best baseline remains remarkably consistent at 9 F1 points across all three tiers, indicating that QASC's advantage is not limited to simple documents but scales proportionally with complexity. This consistency suggests that the query-adaptive mechanism provides a fundamental advantage that is orthogonal to document difficulty.

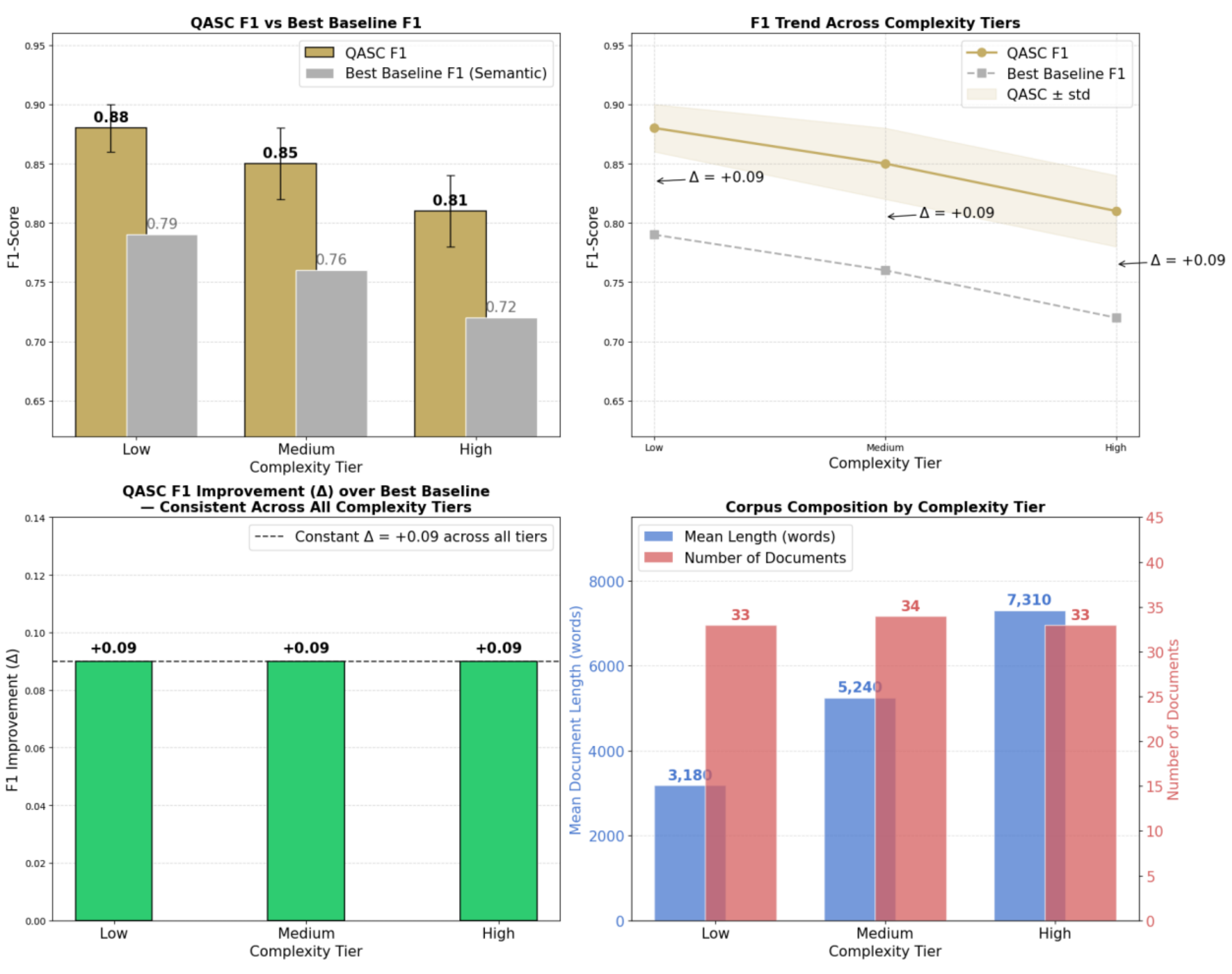

*Figure 4: QASC Performance by Document Complexity Tier*

### 5.5 Human Evaluation Results

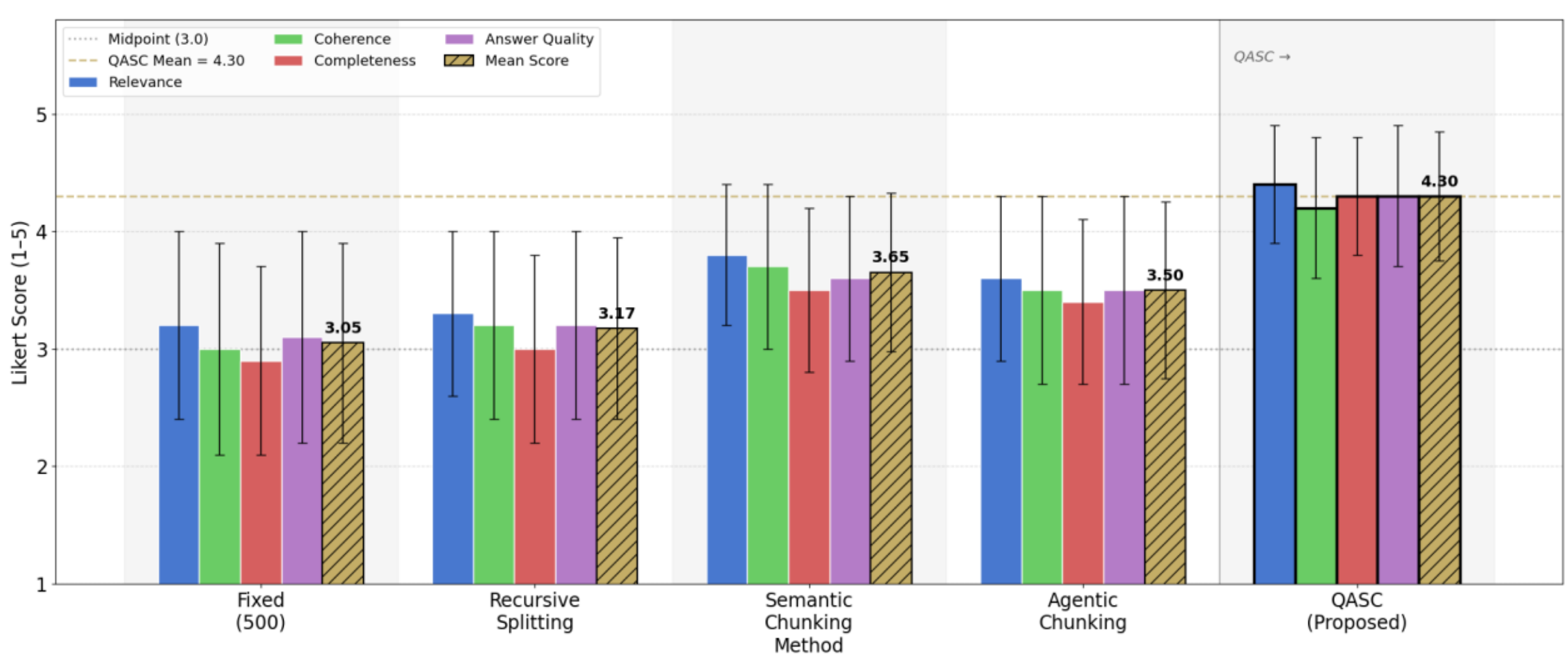

*Figure 5: Human Evaluation Results, 5-Point Likert Scale*

The above Fig 5. Summarizes the human evaluation results. Inter-annotator agreement was strong across all dimensions, with Fleiss' kappa values of 0.78 (Relevance), 0.74 (Coherence), 0.80 (Completeness), and 0.76 (Answer Quality), pairwise Cohen's kappa ranging from 0.72 to 0.84, and disagreements exceeding 2 Likert points occurring in only 3.2% of ratings. QASC received the highest mean ratings across all four dimensions, i.e.  Relevance (4.4), Completeness (4.3), Answer Quality (4.3), and Coherence (4.2), with annotators noting that chunks consistently contained the information needed to answer queries without excessive extraneous material, corroborating the high precision and faithfulness scores from automated evaluation. Coherence ratings further confirm that contextual window expansion and boundary adjustment produce well-formed, readable passages rather than disjointed sentence collections. Annotators identified three recurring strengths: chunks were "focused and on-topic" with minimal irrelevant padding; window expansion made chunks "self-explanatory" without requiring reference to the original document; and generated answers were "more specific and better-supported" than those from baseline methods. The primary criticism was that QASC occasionally missed relevant information located far from any seed sentence, particularly in documents where relevant content was distributed across many small, isolated mentions. This is a limitation that directly motivates future work on non-contiguous passage retrieval

**6. Discussion and Conclusion**

QASC consistently outperforms all baselines across every quality metric (F1 = 0.85 vs. 0.76 for semantic chunking and 0.72 for the best fixed chunking variant), and to understand why it does so, requires examining the information-theoretic dynamics of the RAG pipeline. In standard RAG systems, the retrieval stage must simultaneously identify relevant chunks and hope that relevant information is not diluted by irrelevant co-located material. This is an issue that produces the precision-recall trade-off observed across fixed chunk sizes. QASC resolves this by shifting relevance filtering from the retrieval stage to the chunking stage itself, simplifying the retriever's task from "find the needle in the haystack" to "rank these pre-selected needles," which explains the simultaneous improvement in both precision and recall. This improvement comes at the cost of additional latency: at 380 ms, QASC is 3.5× slower than fixed chunking but remains within the sub-second threshold for interactive applications, and pre-computing sentence embeddings at index time would reduce total latency to approximately 200 ms. A hybrid two-stage architecture which is, query-agnostic pre-chunking for initial retrieval followed by QASC re-chunking on the top-N candidates, shows a promising direction for production deployment. Several concurrent research directions are directly complementary. Proposition-level indexing (Chen et al., 2023) could serve as a seed generation mechanism within QASC's contextual window framework. ColBERT-style token-level interactions (Khattab & Zaharia, 2020) could sharpen seed identification for phrase-sensitive queries and adaptive retrieval methods such as FLARE (Jiang et al., 2023) could be combined with QASC to produce a fully adaptive RAG pipeline where both segmentation and retrieval depth are query-conditioned. In summary, QASC advances document processing for RAG systems by demonstrating that the chunking stage which is often treated as a preprocessing afterthought, is a critical lever for retrieval quality, generation accuracy, and system faithfulness. Evaluated against seven baselines across 40 queries on 100 technical documents, QASC achieves an F1-score of 0.85, representing improvements of 18-27% over fixed chunking, 12% over semantic chunking, and 15% over agentic chunking, alongside a faithfulness score of 0.87 which is the highest among all methods, indicating meaningful reduction in hallucination. Human evaluation (Fleiss' κ = 0.74–0.80) corroborates these findings across all four qualitative dimensions. By integrating the user query directly into segmentation through cosine similarity-based seed selection, contextual window expansion, chunk-level score aggregation, and boundary-aware merging, QASC delivers the right information, in the right context, for the right question. This work therefore establishes a principled foundation for reliable RAG systems across academic research, technical documentation, business intelligence, and healthcare informatics.